# Design optimization of parallel manipulators for high-speed precision machining applications

**Anatol Pashkevich*, Damien Chablat****
**Philippe Wenger*****

**Ecole des Mines de Nantes,*
*Nantes, France, (e-mail: anatol.pashkevich@emn.fr)*
*** Institut de Recherches en Communications et Cybernétique de Nantes,*
*Nantes, France, (e-mail: damien.chablat@irccyn.ec-nantes.fr)*
**** Institut de Recherches en Communications et Cybernétique de Nantes,*
*Nantes, France, (e-mail: philippe.wenger@irccyn.ec-nantes.fr )*

**Abstract:** The paper proposes an integrated approach to the design optimization of parallel manipulators, which is based on the concept of the workspace grid and utilizes the goal-attainment formulation for the global optimization. To combine the non-homogenous design specification, the developed optimization technique transforms all constraints and objectives into similar performance indices related to the maximum size of the prescribed shape workspace. This transformation is based on the dedicated dynamic programming procedures that satisfy computational requirements of modern CAD. Efficiency of the developed technique is demonstrated via two case studies that deal with optimization of the kinematical and stiffness performances for parallel manipulators of the Orthoglide family.

## 1. INTRODUCTION

Many modern material processing operations, especially in automotive and aerospace industry, require high-accuracy positioning and high-speed motion of a work tool that is carried by a robotic manipulator (Brogardh, 2007). However, at present, classical serial manipulators have already reached limits of their performances. In contrast, parallel manipulators are claimed to offer better accuracy, lower mass/inertia properties, and higher structural rigidity (Merlet, 2000). These features are induced by their specific kinematic structure, which eliminates the cantilever-type loading and makes them attractive for innovative machine-tool architectures (Tlusty et al., 1999). But practical utilization of the potential benefits requires development of efficient optimisation techniques, which should satisfy the computational requirement of relevant CAD systems.

Generally, design of parallel manipulators involves simultaneous optimization of two types of criteria that evaluate respectively the kinematic and the kinetostatic/dynamic properties. Both of these groups include a number of performance measures that essentially vary through the workspace but should satisfy the prescribed bounds at any work-point. Up to now, the existing design methods provide separate solutions for the kinematic and kinetostatic objectives, while this paper proposes an integrated approach allowing to achieve desired performances for any prescribed rectangular workspace and to find corresponding parameters taking into account the interaction with the manufacturing process.

In related works, the kinematic design is usually based on the concept of ‘critical points’ that allows essentially reduce the size of the optimization problem and, in some cases, even leads to analytical solutions. In, particular, for the Orthoglide family of manipulators, this approach produced several semi-analytical design strategies that give the manipulators geometry (link lengths and joint limits) for any given cubic workspace with the desired velocity transmission factors. However, the kinetostatic design requires much more computing efforts since it operates with more detailed description of the links (cross-sections, moment of inertia, material properties, etc.). One of a key problem here is computing the stiffness matrix that evaluates the effect of the applied external torques/forces on the compliant displacements of the end-effector.

The main contribution of this paper is in the area of CAD methodology and application of the operation research methods to the integrated design optimization of complex mechanical structures, such as parallel robots. In contrast to the previous work (Pashkevich et al., 2006), the proposed approach does not rely on the ‘critical points’ concept. Instead of this, it operates with the ‘workspace grid’ that is evaluated using a dedicated dynamic-programming-based algorithm allowing, for each particular set of the design parameters, to estimate the largest cuboid-shaped workspace with the desired properties. Further, the workspace parameters are evaluated in the frame of the global optimization, using the goal-attainment formulation. Finally, it yields a set of Pareto-optimal solutions that satisfies the design objectives and are presented the CAD user.

## 2. DESIGN PROBLEM AND METHODOLOGY

Manipulator design traditionally begins with the selection of a kinematic framework and achieving certain geometric goals such as workspace size and dexterity. Besides, for particular manufacturing tasks, the manipulator geometry is optimized with respect to the desired velocity transmission factors. This yields a wire-frame CAD model of a relevant mechanism that defines basic dimensions of the links and spatial locations of all active and passive joints, as well as the joint limits.

Further, when the geometry is established, the design focuses on locating the motion actuators and on the assigning the mass properties (i.e. the links shape, cross-sections, material type, etc.). The primary goal here is the implementation of the above geometric/kinematic objectives while simultaneously satisfying the desired performance requirements (payload, speed, acceleration, accuracy, deflection, etc.). This leads to a complete solid 3D model of the manipulator that is used in the global optimization loop.

Because of very nonlinear and highly coupled relations between the design objectives and the design parameters, the global optimization is still a challenge for the CAD-based design. Existing approaches usually ignore the detailed and physically-reasonable description available in the CAD and operate with the secondary design parameters (masses, moment of inertia, etc.) that are treated as independent ones. The latter may produce a meaningless result that can not be implemented in practice (an evident example can be found in (Asada and Slotine, 1986), where the optimization produced incompatible values of link length, mass and inertia). On the other hand, a straightforward optimization is also non-applicable here because of the very high computing expenses for the CAD- embedded routines that are extensively invoked for the evaluation of the performance measures. These motivate a problem-specific formalization of the manipulator design procedure presented below.

To formulate the design problem, let us define the manipulator geometry by the mapping $\boldsymbol{g}:\mathbf{\Phi}\to\mathbf{W}$, where $\mathbf{\Phi}=\phi_1\times\ldots\phi_n$ and $\mathbf{W}=p_1\times\ldots p_n$ denote respectively the configuration space and the workspace; $\phi_i$ are the joint coordinates, $p_i$ are the coordinates of the end-effector; and $n$ is the number of degrees of freedom. Besides, for each workspace point $\mathbf{p}\in\mathbf{W}$, let us define the matrices $\mathbf{K}_v(\mathbf{p},\boldsymbol{\pi})$, $\mathbf{K}_s(\mathbf{p},\boldsymbol{\pi})$, $\mathbf{K}_m(\mathbf{p},\boldsymbol{\pi})$, that describe various mechanical properties of the manipulator (velocity and force transmission, stiffness, mass distribution, etc.) for any given set of the design parameters $\boldsymbol{\pi}$. Let us also assume that for each type of the matrices $\mathbf{K}_\alpha, \alpha\in\{v,s,m,...\}$, there are defined physically consistent scalar measures $\sigma_\beta(\mathbf{K}_\alpha), \beta\in\{i,t,c,...\}$ that may be directly included in the design objectives or constraints. Some examples of such measures (isotropy, transmission factors, compliance coefficients, etc.) are presented in the following section. Similarly, for the global evaluation of the manipulator, let us introduce the performance measures $\eta_\gamma(\mathbf{g},\boldsymbol{\pi}), \gamma\in\{m,l,w,..\}$ that depend both on the adopted geometrical structure $g$ and the physical parameters of the links $\boldsymbol{\pi}$. Examples of the global measures include the total mass of the manipulator, the length of the principal links, the workspace size, etc.

Then, following the general methodology adopted for the considered application area (high-speed machining), the design optimization problem can be stated as achieving the best values of the performance indices

$$\eta_\gamma(g,\boldsymbol{\pi})\to\min_{\pi}, \quad \forall\gamma \tag{1}$$

subject to the constraints

$$\sigma_\beta\left(\mathbf{K}_\alpha(\mathbf{p},\boldsymbol{\pi})\right)\in S_\beta, \quad \forall\alpha,\beta \;\;\&\;\; \mathbf{p}\in\mathbf{W}_0 \tag{2}$$

that must be satisfied for all point of the cuboid workspace $\mathbf{W}_0$ of size $a\times b\times c$, which includes the manufacturing task. It should be noted that the latter assumption (concerning the workspace shape) is essential here and allows considerably speed-up the optimization routines. Since in practice this problem can not be solved by the direct search methods, in the following subsections there will be presented the discretisation scheme and relevant optimization algorithms allowing to obtain desired solutions in reasonable time.

## 3. PERFORMANCE MEASURES

Let us present the most essential performance measures that are used in mechanical design of manipulators. Traditionally they are directly included in the design constraints/objectives to be satisfied or optimized throughout the prescribed workspace. However, in this paper each performance measure is preliminary converted in an alternative form that defines the workspace subset where the relevant criterion is higher/lower of the desired value.

### *3.1 Geometric and Kinematic Performances*

In robot design, the manipulator architecture is usually defined outside of the main optimization loop and highly depends on an application target. Within the CAD system, this architecture is described as an assembly of the links/joints that is parameterized by the link lengths and the joint limits. This set of the parameters is sufficient for evaluating the basic geometric and kinematic specifications such as the workspace size, dexterity, velocity transmission, reachability, etc. Using this model, the workspace $\mathbf{W}$ may be generated in a straightforward way, using the direct kinematic equations $\boldsymbol{p}=\boldsymbol{g}(\boldsymbol{\phi})$ and the joint limits $\mathbf{\Phi}$. It worth mentioning that, for parallel robots, the direct kinematics is usually non-trivial and an analytical solution do not exist in the general case (Merlet, 2006).

Since for the considered application (high-speed machining) the desired workspace is the cuboid $\mathbf{W}_0$ of size $\{a\times b\times c\}$, the relevant performance measure may be defined by the largest similar object $\{\mu a\times\mu b\times\mu c\}$ inscribed in $\mathbf{W}_0$, i.e.

$$\mathbf{W}^{abc}=\mathbf{T}(\mu\mathbf{W}_0); \quad (\mu,\mathbf{T})=\arg\max_{\mu,\mathbf{T}}\left\{\mu \;\middle|\; \mathbf{T}(\mu\mathbf{W}_0)\subset\mathbf{W}\right\} \tag{3}$$

where μ, **T** are respectively the scalar scaling factor and the coordinate transformation operator in the Cartesian space. This notion is the fundamental issue of this paper and is discussed in details in the following sections.

For the kinematic performances, complete information is contained in the Jacobian matrix $\mathbf{J} = \partial\mathbf{g}(\phi)/\partial\phi$ that is commonly evaluated using the condition number $cond(\mathbf{J})$, or the largest/smallest singular values $\sigma_{\min}, \sigma_{\max}$ (the latter are also referred to as the "velocity transmission factors"). However, because the Jacobian varies throughout the workspace, it should be defined a global metric that is usually computed by averaging or by detecting the worst case. Hence, in this paper, we suggest to redefine the global kinematic metric by using the above notion of the "largest inscribed cuboid". For instance, if the condition number $k(\mathbf{J})$ is used as the local metric, the manipulator global performances is evaluated by the size of the cuboid workspace that satisfies the design specification $k(\mathbf{J}(\mathbf{p})) \le k_{\max}$ :

$$\mathbf{W}_k^{abc} = \mathbf{T}(\mu\mathbf{W}_0); \tag{4}$$

$$(\mu,\mathbf{T}) = \arg\max_{\mu,\mathbf{T}}\left\{\ \mu \ \middle|\ k\left(\mathbf{J}(\mathbf{p},\boldsymbol{\pi})\right) \le k_{\max};\ \mathbf{p} \in \mathbf{T}(\mu\mathbf{W}_0) \subset \mathbf{W}\right\}$$

This allows operate in a similar way with both geometric and kinematic performance measures.

### 3.2 Elastic Performances

For parallel manipulators, elasticity is an essential performance measure since it is directly related to the positioning accuracy and the payload capability. Mathematically, this benchmark is defined by the stiffness matrix, which describes the relation between the linear/angular displacements of the end-effector (wrench) and the external forces/torques applied to the tool. It is obvious that the elasticity is highly dependent upon geometry, materials and link shapes that are completely defined within the CAD model.

The stiffness matrix may be computed using three different methods: the Finite Element Analysis (FEA), the matrix structural analysis (SMA), and the virtual joint method (VJM) (Alici & Shirinzadeh, 2005). The first of them, FEA, is proved to be the most accurate and reliable but requires very high computational efforts for repeated 3D remeshing over the whole workspace. The second method, SMA, also incorporates the main ideas of the FEA, but operates with rather large structural elements that allow some reduction of the computational expenses. And finally, the VJM method is based on the expansion of the traditional rigid model by adding the virtual joints (localized springs), which describe the elastic deformations of the links. It is the most efficient technique for the design optimization, which was recently enhanced by the authors to handle a case of the overconstrained manipulators (Pashkevich et al., 2008).

Within this approach, each *i*th kinematic chain of the manipulator is described by the kinetostatic model that defines the differential kinematics and elasticity taking into account the active, passive and virtual joints:

$$\delta\mathbf{t}_i = \mathbf{J}_\theta^i \cdot \delta\boldsymbol{\theta}_i + \mathbf{J}_q^i \cdot \delta\mathbf{q}_i$$
$$\boldsymbol{\tau}_\theta^i = \mathbf{K}_\theta \cdot \delta\boldsymbol{\theta}_i,\quad \boldsymbol{\tau}_q^i = \mathbf{0},\quad i = 1,\dots n \tag{5}$$

where vector $\delta\mathbf{t}_i$ describes the end-effector translation and rotation (wrench) with respect to the Cartesian axes; vector $\delta\boldsymbol{\theta}_i$ collects all virtual and active joint coordinates, vector $\delta\mathbf{q}_i$ includes all passive joint coordinates, $\mathbf{J}_\theta^i$ and $\mathbf{J}_q^i$ are the kinematic Jacobians with respect to the virtual/actuated and passive joints, $\boldsymbol{\tau}_\theta^i$, $\boldsymbol{\tau}_q^i$ are the aggregated vectors of the virtual/active and passive joint reactions, and $\mathbf{K}_\theta$ is the aggregated spring stiffness matrix of the links (composed of 6x6 stiffness matrices of all virtual springs and the actuator stiffness coefficients). As it proved in (Pashkevich et al., 2008), the desired force-wrench relation $\mathbf{f}_i = \mathbf{K}_i \cdot \delta\mathbf{t}_i$ can be computed from the system

$$\begin{bmatrix} \mathbf{J}_\theta^i\,\mathbf{K}_\theta^{-1}\,\mathbf{J}_\theta^{iT} & \mathbf{J}_q^i \\ \mathbf{J}_q^{i\,T} & \mathbf{0} \end{bmatrix} \cdot \begin{bmatrix} \mathbf{f}_i \\ \delta\mathbf{q}_i \end{bmatrix} = \begin{bmatrix} \delta\mathbf{t}_i \\ \mathbf{0} \end{bmatrix}, \tag{6}$$

by the straightforward inversion of the relevant matrix and extracting from it the upper-left sub-matrix of size 6×6. Then, for the entire manipulator, the stiffness is computed by aggregation all kinematic chains: $\mathbf{K}_m = \sum_{i=1}^{n} \mathbf{K}_i$

Using the stiffness matrix, one can evaluate the elastic deflection at the tool-point caused by a particular machining operation. However, following the above proposed approach, it is prudent to compute a relevant workspace-based metric, i.e.

$$\mathbf{W}_s^{abc} = \mathbf{T}(\mu\mathbf{W}_0); \tag{7}$$

$$(\mu,\mathbf{T}) = \arg\max_{\mu,\mathbf{T}}\left\{\ \mu \ \middle|\ \left\|K_m\left(\mathbf{p},\boldsymbol{\pi}\right)\cdot\mathbf{f}_m\right\| \le \varepsilon_{\max};\ \mathbf{p} \in \mathbf{T}(\mu\mathbf{W}_0) \subset \mathbf{W}\right\}$$

where $\mathbf{f}_m$, $\varepsilon_m$ are respectively the eternal force and the upper limit of the elastic deflection defined by the design specifications.

### 3.3 Dynamic Performances

The manipulators dynamics is determined by the link inertial characteristics that bound the highest reachable accelerations and define capability to execute a given manufacturing task. Relevant evaluation techniques are generally based on two concepts: the dynamic isotropy and the dynamic manipulability. The first of them (Asada, 1986) deals with generalized inertia ellipsoid (GIE) that geometrically represents the generalized inertia matrix $\mathbf{G} = \mathbf{J}^{-T}\mathbf{D}\,\mathbf{J}^{-1}$ referring to the end effector, where $\mathbf{D}(\mathbf{q},\boldsymbol{\pi})$ is the manipulator inertia matrix with respect to the joint space and $\mathbf{J}(\mathbf{q},\boldsymbol{\pi})$ is the kinematic Jacobian. Using the GIE, the design is aimed at transforming the ellipsoid to a sphere (without respect to its radius), in order to ensure that the attitude to produce end-effector accelerations does not depend on the direction. Hence, from practical point view, this classical approach seeking for $cond\left(\mathbf{D}(\mathbf{p})\right)=1$ should be reformulated taking into account the GEI size. The latter can be done by bounding the norm $\|\mathbf{D}(\mathbf{p})\| \le m_0$ , for instance the spectral norm allows to bound the GEI axis length. Further, similarly to the above, it is necessary to compute a relevant workspace-inscribed cuboid, i.e.

$$\mathbf{W}_m^{abc} = \mathbf{T}(\mu \mathbf{W}_0); \qquad (8)$$

$$(\mu, \mathbf{T}) = \arg\max_{\mu,\mathbf{T}} \left\{ \mu \;\middle|\; \|\mathbf{G}(\mathbf{p},\boldsymbol{\pi})\| \le m_0;\; \mathbf{p} \in \mathbf{T}(\mu \mathbf{W}_0) \subset \mathbf{W} \right\}$$

Similar approach can be applied to the dynamic manipulability metric (Yoshikawa, 1985) that evaluates ability of the manipulator to transform the inputs forces/torques into the output accelerations. This transformation is defined by the matrix product $\mathbf{J}(\mathbf{p},\boldsymbol{\pi}) \cdot \mathbf{D}(\mathbf{p},\boldsymbol{\pi})^{-1}$, so the corresponding cuboid is defined as

$$\mathbf{W}_a^{abc} = \mathbf{T}(\mu \mathbf{W}_0); \qquad (9)$$

$$(\mu, \mathbf{T}) = \arg\max_{\mu,\mathbf{T}} \left\{ \begin{array}{c} \mu \;\mid\; \|\mathbf{J}(\mathbf{p},\boldsymbol{\pi}) \cdot \mathbf{D}(\mathbf{p},\boldsymbol{\pi})^{-1} \cdot \boldsymbol{\tau}\| \ge a_{\min}; \\ \mathbf{p} \in \mathbf{T}(\mu \mathbf{W}_0) \subset \mathbf{W}; |\tau_i| \le \tau_i^{\max} \end{array} \right\}$$

where $a_{\min}$, $\tau_i^{\max}$ are the desired acceleration and the maximum force/torque in the ith actuated joint respectively.

Thus, independent of the physical meaning, all performance measures are presented in a similar way allowing essentially simplify the design optimization process.

## 4. WORKSPACE-BASED METRICS

The above presented technique for the global evaluation of the manipulator performances is based on an auxiliary computational problem, i.e. estimation of the largest cuboid-shaped sub-workspace where the relevant criterion is higher or lower of the desired value. Let us attack this problem numerically, using the workspace discretisation and applying the dynamic programming.

To satisfy the desired cuboid proportions $\{a \times b \times c\}$, let us define the workspace grid $\{G_{ijk}\}$ that includes the manipulator workspace $W_0$ and possesses uniform but different steps along the Cartesian axes, namely $(a/N_0;\ b/N_0;\ c/N_0)$, where $N_0$ defines the discretisation precision. Besides, for each node of the grid, let us compute relevant local performance measure and define a 3D binary matrix $\Omega_{ijk} \in \{0,1\}$, where $\Omega_{ijk}=1$ if the corresponding design constraint/objective is satisfied, and $\Omega_{ijk}=0$ otherwise. For computation conveniences, let us also set $\Omega_{ijk}=0$ if $G_{ijk} \notin W_0$.

Thus, the original problem is reduced to searching for the largest cubic submatrix inside of $\{\Omega_{ijk}\}$ containing non-zero values only. The latter can be efficiently solved applying the following algorithm that operates with additional integer matrix $\{\Phi_{ijk}\}$ that define sizes of the candidate solutions with the vertexes $(i,j,k)$:

---

**Step 0**. *Set* $\Phi_{ijl}=0, \quad \forall i,j,k$

**Step 1**. *Set* $\Phi_{ijk}=\Omega_{ijk}$ for
$\{i=1\,\&\forall j,k\} \cup \{j=1\,\&\forall i,k\} \cup \{k=1\,\&\forall i,j\}$

**Step 2**. *for* $i=2:i_{\max}$ *do*
*for* $j=2:j_{\max}$ *do*
*for* $k=2:k_{\max}$ *do*
*if* $\Omega_{ijk}=1$ *then*

$$\Phi_{ijk} = 1 + \min \left\{ \begin{array}{l} \Phi_{i-1,j,k},\ \Phi_{i,j-1,k},\ \Phi_{i,j,k-1}, \\ \Phi_{i-1,j-1,k},\ \Phi_{i-1,j,k-1},\ \Phi_{i,j-1,k-1}, \\ \Phi_{i-1,j-1,k-1} \end{array} \right\}$$

**Step 3**. *Find* $d=\max(\Phi_{ijk})-1$ ; $(i_0,j_0,k_0)=\arg\max(\Phi_{ijk})$

Step 4. *Retrieve* from the grid $\{G_{ijk}\}$ the desired cuboid bounded by the indices $(i_0-d, j_0-d, k_0-d)$ and $(i_0,j_0,k_0)$.

---

Validity of this routine and correctness of the relevant recurrent expression can be proved using the standard ideas of the dynamic programming, similar to finding the largest square block in two-dimensional binary matrix.

Hence, for each performance measure and each set of the design parameters, it can be computed a workspace-based metrics composed of the coordinate ranges of the largest cuboid-shaped sub-workspace with the desired proportions.

## 5. DESIGN PROCEDURE

Following the common engineering concepts adopted in robotic practice, let us divide the design procedure into two basic steps: (i) *geometric/kinematic design* that provides the manipulator geometry, including the link lengths, joint locations and joint limits, and (ii) integrated *kinetostatic/dynamic design* that deals with assigning the shape and mass properties for all joints and links. This separation allows essentially simplify optimization algorithms and improve their convergence.

In the frame of the proposed methodology, both design problems can be reformulated in a similar way:

$$f_i(\boldsymbol{\pi}) \to \min_{\boldsymbol{\pi}}, \quad \forall i \qquad (10)$$

subject to the workspace-based constraints

$$size(W_j^{abc}(\boldsymbol{\pi})) \ge \{a_0 \times b_0 \times c_0\} \quad \forall j \qquad (11)$$

where indices $\alpha$ , $\beta$ define particular design objectives and constraints. For instance, for the geometric and kinematic design, the objectives are to minimize the manipulator dimensions (link lengths), while the constraints define the desired workspace size and the range of the velocity transmission factors. For the kinetostatic and dynamic design, the objectives deal with minimizing the component masses, and the constraints identify the prescribed stiffness and the accelerations transmission characteristics (for the fixed geometry).

For computational conveniences, let us transform the design constraints and present them in the scalar form as $h_k(\boldsymbol{\pi}) \geq h_k^0$. Then, to generate a specific Pareto-optimal solution, let us apply the goal attainment technique that yields the following nonlinear programming formulation:

$$\lambda \to \min_{\lambda,\boldsymbol{\pi}} \qquad (12)$$

subject to

$$f_i(\boldsymbol{\pi}) - w_i\lambda \leq f_i^0; \quad h(\boldsymbol{\pi}) \geq h_i^0; \ \forall i \qquad (13)$$

Here $\lambda$ is an unrestricted scalar variable, $w_i \geq 0$ are designer-selected weighting coefficients, and $f_i^0$ are the goals to be realized for each design objective (usually, the designer can extract them from the design specifications). In this formulation, minimization of λ, tends to force the specifications to meet their goal. If, at the solution point, λ is negative, the goals have been over-attained; if λ is positive, then the goals have been under-attained. The method is appealing since it is possible for the designer to specify unrealizable objectives and still obtain a solution which represents a compromise. Advantages of this approach will be demonstrated in the following section that focuses on the application examples.

## 6. APPLICATION EXAMPLES

### *6.1 Optimization of Orthoglide Geometry*

Let us consider first the problem of the geometric synthesis of the Orthoglide parallel manipulators that was addressed in our previous paper (Pashkevich et al., 2006). However, the previous solution is valid only for the "cubic type" design specifications (i.e. for the case $a_0 = b_0 = c_0$). Hence, it is prudent to generalize the former results for a general case.

The Orthoglide architecture includes three identical parallel chains that are actuated by linear drives with mutually orthogonal axes. Each kinematic chain is formally described as PRPaR, where P, R and Pa denote the prismatic, revolute, and parallelogram joints respectively. The output machinery is connected to the legs in such a manner that the tool moves in the Cartesian space with fixed orientation (i.e. restricted to translational motions). Assuming that the manipulator legs are not necessarily equal, the manipulator Jacobian may be described by the following equations

$$\mathbf{J}_\rho = \begin{bmatrix} 1 & \dfrac{p_y}{p_x-\rho_x} & \dfrac{p_z}{p_x-\rho_x} \\ \dfrac{p_x}{p_y-\rho_y} & 1 & \dfrac{p_z}{p_y-\rho_y} \\ \dfrac{p_x}{p_z-\rho_z} & \dfrac{p_y}{p_z-\rho_z} & 1 \end{bmatrix}^{-1} \qquad (14)$$

where $\mathbf{p} = (p_x, p_y, p_z)^T$ is the output vector of the TCP position, $\boldsymbol{\rho} = (\rho_x, \rho_y, \rho_z)^T$ is the input vector of the prismatic joints variables, $i \in \{x, y, z\}$. It allows computing the condition number and the velocity transmission factors for each workspace point $(p_x, p_y, p_z)$. Then, using the workspace grid, each set of the geometrical parameters $\{L_x, L_y, L_z\}$ can be evaluated by the metric $\{a \times b \times c\}$.

The optimization results were normalized with respect to the desired workspace size and tabulated. One of the case studies corresponding to the design specifications $W_0 = \{1.0 \times 1.0 \times 0.8\}$ is presented in Fig 1. It contains solutions constrained by the equality $L_1 = L_2$ incited by the shape of $W_0$. As follows from them, the developed technique perfectly accounts non-symmetry of the design specifications and produces relevant non-symmetrical solutions for the manipulator geometry. For example, compared to the known solution corresponding to the symmetrical case, the obtained Pareto-optimal set contains solutions, which allow reduce the manipulator dimensions by 10…20 %.

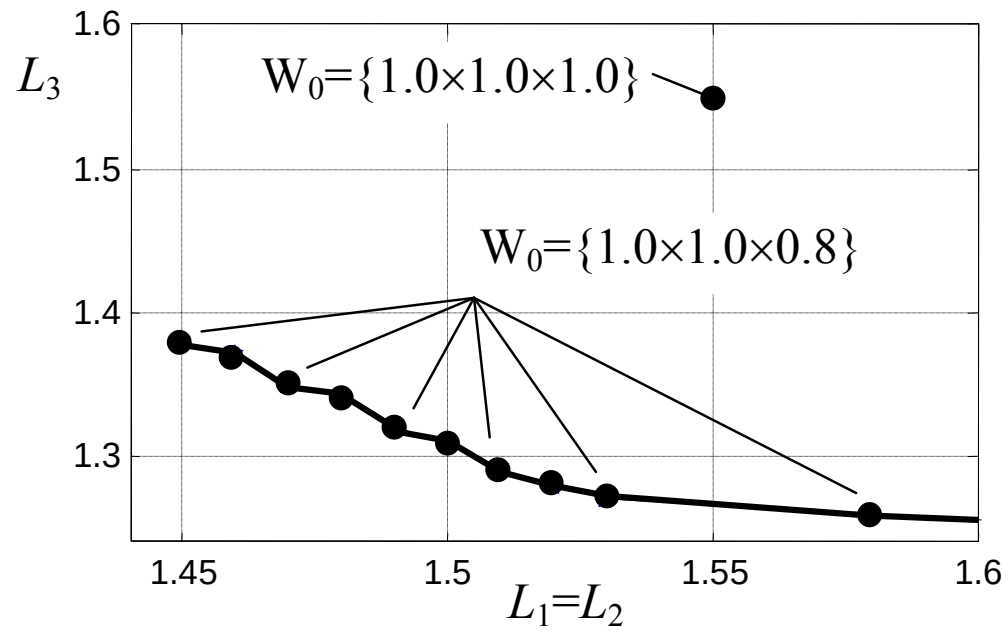

Fig. 1. Optimal geometrical parameters of Orthoglide.

### *6.2 Optimization of Orthoglide Stiffness*

The stiffness-based design assumes that the manipulator geometry is constrained and satisfies the kinematic objectives. Hence, before considering a related optimization problem, it is prudent to investigate the relationship between the kinematic and stiffness specifications for a particular manipulator, such as Orthoglide prototype with a symmetrical architecture (Chablat & Wenger, 2003). The kinematic specifications for this manipulator were defined via the velocity transmission factors that should be in the range of 0.5…2.0 for the workspace size 200×200×200 mm$^3$.

The manipulator stiffness model was derived using the "virtual-joint" technique, and the model parameters were evaluated via the FEA. Then, the workspace was meshed with a step of 1.0 mm and the stiffness matrices were computed for each node. Using these data and applying the proposed algorithm, it was generated a set of nested cubic subspaces with the upper-bounded translational/rotational stiffness. Corresponding results demonstrate good agreement between the shapes of the velocity and stiffness maps. In particular, the location of the cubic workspaces derived via constraining the velocity transmission factors and the stiffness are almost the same (difference is less than 5%). Also, the obtained results justify the velocity transmission constraint [0.5; 2.0] used in our previous study, since its violation leads to essential decrease of the stiffness.

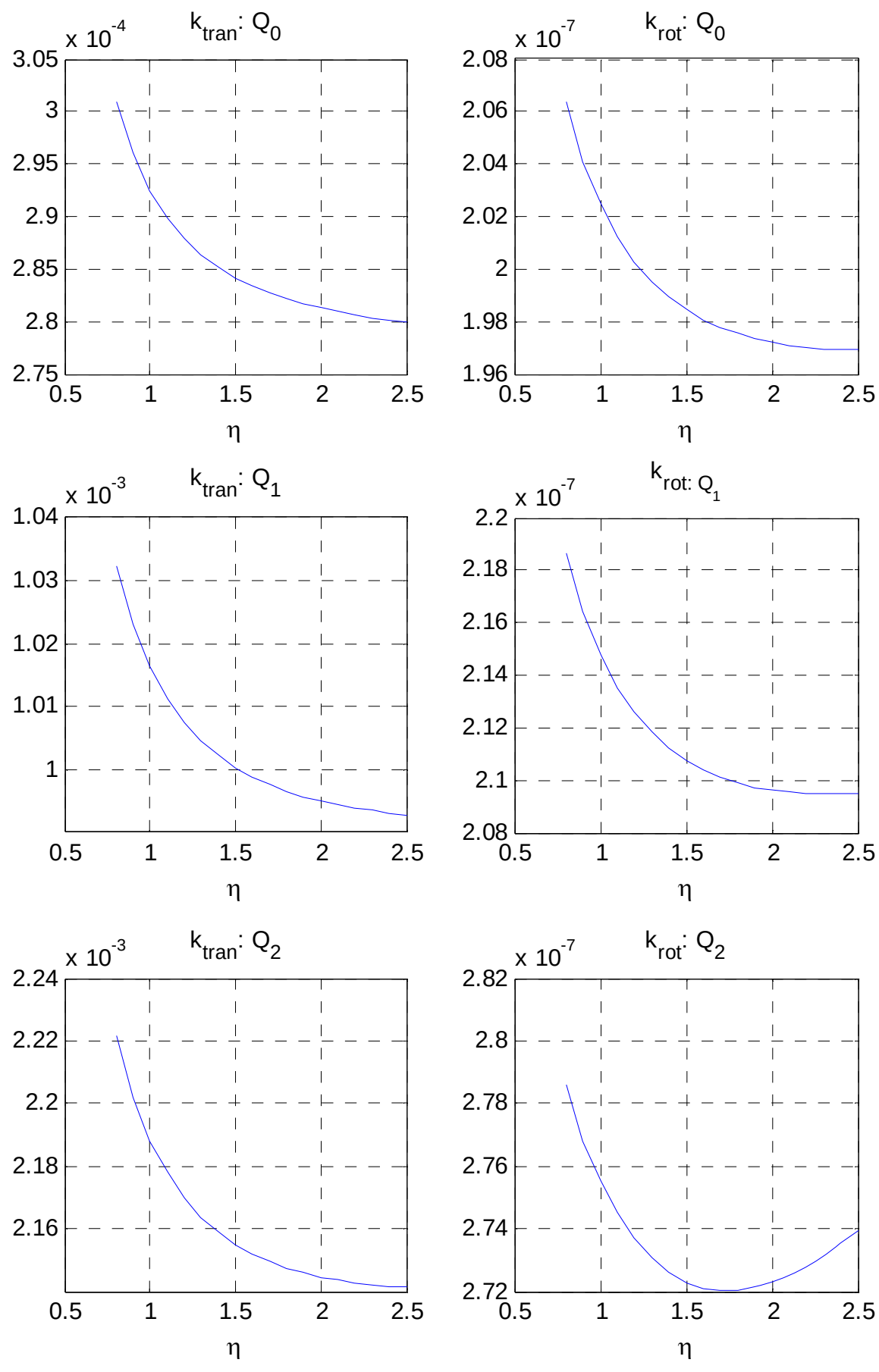

Fig. 2. Optimisation of the Orthoglide parallelogram links

At the second stage, the developed technique was applied to parametrical optimization of the Orthoglide components with respect to the stiffness objectives. Here, it was assumed that the link lengths are constrained while the cross-sectional dimensions were scaled using the factors $\mu$ and $1/\mu$. The latter allowed maintaining the same cross section areas and, respectively, the link masses.

An extraction from the optimization results concerning the parallelogram links are presented in Fig. 2. For the sake of clarity, the stiffness is evaluated in three characteristic points $Q_0$, $Q_1$ and $Q_2$, where $Q_0$ corresponds to the isotropic posture and $Q_1$, $Q_2$ define the workspace boundary. As follows from them, the manipulator stiffness can be essentially improved without increasing the link masses, just by changing the cross-section shape according to the scaling factor $\mu \approx 1.8$. These results have been also used for the design of a new Orthoglide prototype, which is currently under development in our laboratory.

## 7. CONCLUSIONS

Parallel robotic manipulators present attractive solutions for innovative machine-tool architectures, which should insure high-speed and precision machining of large spectrum of materials. However, practical utilization of the potential benefits requires development of efficient optimisation techniques, which should satisfy the computational speed and accuracy requirements of relevant CAD procedures. To response this challenge, the paper proposes an integrated approach to the design optimization of parallel manipulators. In contrast to previous works, which are based on the concept of "critical points", the proposed technique is compatible with current CAD methodology and allows achieving desired performances for any prescribed workspace and to take into account an interaction with the manufacturing process.

The developed approach has been validated by a number of case studies, which focus on the design optimization of translational parallel manipulators of the Orthoglide family. Further work will deal with extending these results for more general case, including Orthoglide with additional orientational axes that is currently under development.

## REFERENCES

Alici, G., Shirinzadeh, B. (2005). Enhanced Stiffness Modeling, Identification and Characterization for Robot Manipulators. *IEEE Transactions on Robotics*, **21**(4), 554 – 564.

Asada, H., Slotine. J. (1986). *Robot Analysis and Control*. Wiley, New York.

Brogardh, T. (2007). Present and future robot control development - An industrial perspective. *Annual Reviews in Control*, **31**(1), 69-79.

Chablat D., Wenger Ph. (2003). Architecture Optimization of a 3-DOF Parallel Mechanism for Machining Applications, the Orthoglide. I*EEE Transactions on Robotics and Automation*, 19(3), pp. 403-410.

Chanal, H., Duc, E., Ray, P. (2006). A study of the impact of machine tool structure on machining processes. *International Journal of Machine Tools and Manufacture*, **46**(2), 98-106.

Chen, S.F., Kao, I. (2000). Conservative congruence transformation for joint and Cartesian stiffness matrices of robotic hands and fingers. *International Journal of Robotics Research*, **19**(9), 835-847.

Griffis, M., Duffy, J. (1993). Global stiffness modeling of a class of simple compliant couplings, *Mechanism and Machine Theory*, **28**(2), 207–224.

Li, Y., Xu, Q. (2008). Stiffness analysis for a 3-PUU parallel kinematic machine. *Mechanism and Machine Theory*, 43(2), 186-200.

Merlet, J.-P. (2006). *Parallel Robots*, Kluwer Academic Publishers, Dordrecht.

Pashkevich A., Chablat D., Wenger Ph. (2006) Kinematics and Workspace Analysis of a Three-Axis Parallel Manipulator: the Orthoglide, *Robotica*, 24(1), 39-49.

Pashkevich A., Chablat D., Wenger Ph. (2009). Stiffness analysis of overconstrained parallel manipulators. *Mechanism and Machine Theory*, **44**(5), 966-982. Available online 25 July 2008.

Tlusty J., Ziegert J.C., Ridgeway S. (1999). Fundamental comparison of the use of serial and parallel kinematics for machine tools, *CIRP Annals*, 48(1), pp. 351–356.

Yoshikawa T. (1985) Dynamic manipulability of robot manipulators. *Journal of robotic systems*. 2(1), pp. 113-124.